\pdfoutput=1

\documentclass[runningheads]{llncs}
\usepackage[width=122mm,left=12mm,paperwidth=146mm,height=193mm,top=12mm,paperheight=217mm]{geometry}
\def\IEEEConf{0}
\ifx\CSMacrosIncluded\undefined         % This ``if'' ensures that this file is included only once
\def\CSMacrosIncluded{1}

\usepackage[]{csquotes}                 % Error-checking and flexible quotation markup
\usepackage[]{bigfoot}                  % Everything related to footnotes
\usepackage[]{enumitem}                 % Everything one could want for formatting lists
\usepackage[]{fixltx2e}                 % Fixes and adds many little details. Better to be safe than sorry
\usepackage[]{microtype}                % Enables PdfTeX's microtypographical features: margin
\usepackage[]{textpos}                  % For putting things anywhere on the page, even using
\usepackage[]{amsmath}                  % Handy tools for mathematical typesetting
\usepackage[]{mathtools}                % Mathtools package requires and fixes amsmath
\usepackage[]{amsthm}
\usepackage[]{amssymb,amsfonts}
\usepackage{graphicx}
\usepackage[]{bm}

\usepackage[]{svn-multi}
\usepackage[]{xcolor}
\usepackage[mathscr]{eucal}

\usepackage[]{xparse}
\usepackage[]{xstring}

\usepackage[]{xspace}
\usepackage[]{comment}
\usepackage{enumitem}                   % Gives more powerful enumeration environments
\usepackage[bookmarks=true,bookmarksopen=true,
            breaklinks=true,
            colorlinks=true,
            urlcolor=blue,linkcolor=blue,filecolor=blue,citecolor=blue,
            menucolor=blue,
            pdfpagelabels=true,hypertexnames=true,
            plainpages=false,naturalnames=false]{hyperref}

\usepackage{algpseudocode}              % Example:
\newcommand\foreign[1]{\emph{#1}}

\makeatletter
\newcommand\ensurecomma{\@ifnextchar,{}{\@latex@error{Don't forget the comma!}{}}}
\makeatother

\newcommand\ie{\foreign{i.e.}\ensurecomma}

\makeatletter
\newcommand\ensuresingleperiod{\@ifnextchar.{}{.\xspace}}
\makeatother

\newcommand\etc{\foreign{etc}\ensuresingleperiod}
\newcommand\etal{\foreign{et al}\ensuresingleperiod}

\usepackage{expl3,xparse}               % These two are LaTeX3 packages
\ExplSyntaxOn
\cs_new_eq:Nc \emph_old:n { emph~ }     % Copying the old definition of `\emph`
\cs_new_eq:NN \emph_braces:n \textup    % Set up how braces should be typeset.
\cs_new:Npn \emph_new:n #1 {
  \tl_set:Nn \l_emph_tl {#1}
  \tl_replace_all:Nnn \l_emph_tl {(}{\emph_braces:n{(}}
  \tl_replace_all:Nnn \l_emph_tl {)}{\emph_braces:n{)}}
  \tl_replace_all:Nnn \l_emph_tl {[}{\emph_braces:n{[}}
  \tl_replace_all:Nnn \l_emph_tl {]}{\emph_braces:n{]}}
  \exp_args:NV \emph_old:n \l_emph_tl
}
\RenewDocumentCommand {\emph} {sm} {
  \IfBooleanTF {#1} {\emph_old:n {#2}} {\emph_new:n {#2}}
}
\ExplSyntaxOff

\def\inlinegraph#1{{\setbox0=\hbox{\includegraphics[height=40mm]{#1}}
    \dimen0=0.7ex \advance\dimen0 by -0.5\ht0 \raise\dimen0\box0}}

\let\latexcirc=\circ
\newcommand{\ccirc}{\mathbin{\mathchoice
  {\xcirc\scriptstyle}
  {\xcirc\scriptstyle}
  {\xcirc\scriptscriptstyle}
  {\xcirc\scriptscriptstyle}
}}
\newcommand{\xcirc}[1]{\vcenter{\hbox{$#1\latexcirc$}}}
\let\circ\ccirc
\renewcommand{\vec}[1]{\ensuremath{\pmb{#1}}}
\newcommand{\mat}[1]{\ensuremath{\mathbf{#1}}}
\newcommand{\set}[1]{\ensuremath{\mathscr{#1}}}

\makeatletter
\@tfor\next:=abcdefghijklmnopqrstuvwxyxz\do
 {\begingroup\edef\x{\endgroup
    \noexpand\@namedef{v\next}{\noexpand\vec{\next}}%
  }\x}
\@tfor\next:=ABCDEFGHIJKLMNOPQRSTUVWXYZ\do
 {\begingroup\edef\x{\endgroup
    \noexpand\@namedef{m\next}{\noexpand\mat{\next}}%
  }\x}
\@tfor\next:=ABCDEFGHIJKLMNOPQRSTUVWXYZ\do
 {\begingroup\edef\x{\endgroup
    \noexpand\@namedef{s\next}{\noexpand\set{\next}}%
  }\x}
\makeatother

\let\T\Transpose

\excludecomment{fixme}

\noexpandarg

\noexpandarg

\noexpandarg
\DeclareDocumentCommand{\inprod}{ O{,} m }{%
\StrBefore{#2}{#1}[\leftmat]%
\StrBehind{#2}{#1}[\rightmat]%
{\ensuremath{  \left\langle \leftmat, \rightmat \right\rangle }}%
}

\ifdefined\IEEEConf
\makeatletter
\newdimen\bibindent
\newdimen\bibspacing
\makeatother
\usepackage[margin=2pt,font=scriptsize,labelfont=bf, labelsep=endash]{caption}

\else                                  % \IEEEConf not defined
\fi                                     % ------------------  \IEEEConf  ---------------------
\else
\fi
\graphicspath{{./}{./Fig/}{./Figs/}{./Fig/eps/}{./Fig/pdf/}}

\usepackage{amsmath}
\usepackage{url}
\usepackage{multirow}
\usepackage{subfig}

\renewcommand{\paragraph}{\textbf}

\newcommand{\revised}[1]{{#1}}

\def\FPR{{\sc FPR}\xspace}

\def\pAUC{{\rm pAUC}\xspace}
\def\paucstruct{{\rm pAUC$^{\,\rm struct}$}\xspace}

\def\vxip{\vx_i^{+}}

\def\vxjfzm{\vx_{(j)_{f|\zeta}}^{-}}

\def\jalpha{j_\alpha}
\def\jbeta{j_\beta}

\def\HOGSVM{\texttt{HOG+SVM}\xspace}
\def\HOGBOOST{\texttt{HOG+BOOSTING}\xspace}

\def\SCOV{{\rm sp-Cov}\xspace} %
\def\nSCOV{{\rm sp-Cov}}
\def\SLBP{{\rm sp-LBP}\xspace}
\def\nSLBP{{\rm sp-LBP}}
\def\ChnFtrs{\texttt{ChnFtrs}\xspace}

\def\ACF{\texttt{ACF}\xspace}
\def\nACF{\texttt{ACF}}

\def\v1{{\boldsymbol 1}}

\def\atan2{{\texttt{atan2}\xspace}}

\begin{document}
\pagestyle{headings}
\mainmatter
\title{Strengthening the Effectiveness of Pedestrian \\
Detection with Spatially Pooled Features\thanks{Appearing in Proc.\ European Conf. Computer Vision, 2014.}}

\titlerunning{Strengthening the Effectiveness of Pedestrian Detection
}

\authorrunning{Paisitkriangkrai, Shen, van den Hengel}

\author{Sakrapee Paisitkriangkrai, Chunhua Shen\thanks{Corresponding author:
(email: chunhua.shen@adelaide.edu.au).}, Anton van den Hengel}
\institute{The University of Adelaide, Australia}

\maketitle

\begin{abstract}

We propose a simple yet effective approach to the problem of pedestrian detection which outperforms the current state-of-the-art.
Our new features are built on the basis of low-level visual features and spatial pooling.
Incorporating spatial pooling improves the translational invariance
and thus the robustness of the detection process.
We then directly optimise the partial area under the ROC curve (\pAUC) measure, which concentrates detection performance in the range of most practical importance.
The combination of these factors leads to a pedestrian detector which outperforms all competitors on all of the standard benchmark datasets.
We advance state-of-the-art results by lowering the average miss rate from $13\%$ to $11\%$ on
the INRIA benchmark, $41\%$ to $37\%$ on the ETH benchmark,
$51\%$ to $42\%$ on the TUD-Brussels benchmark and $36\%$ to $29\%$ on the
Caltech-USA benchmark.

\end{abstract}

\section{Introduction}

Pedestrian detection is a challenging but an important problem
due to its practical use in many computer vision applications
such as video surveillance, robotics and human computer interaction.
The problem is
made difficult by the inevitable variation in target
appearance, lighting and pose, and by occlusion.
In a recent literature survey on pedestrian detection \cite{Dollar2012Pedestrian}
the authors evaluated several pedestrian detectors and
concluded that combining multiple features
can significantly boost the performance of pedestrian detection.

Hand-crafted low-level visual features have been applied to several computer vision
applications and shown promising results
\cite{Dalal2005HOG,FisherBoost2013IJCV,paisitkriangkrai2008fast,Tuzel2008Pedestrian,Viola2004Robust,Wang2009HOG}.
Inspired by the recent success of spatial pooling on object recognition
and pedestrian detection problems
\cite{Park2013Exploring,Sermanet2013Pedestrian,Wang2013Regionlets,Yang2009Linear},
we propose to perform the spatial pooling operation to create the new feature type.
Our new detector yields competitive results to the state-of-the-art
on major benchmark data sets.
A further improvement is achieved when
we combine the new feature
type and
channel features from \cite{Dollar2014Fast}.
We confirm the observation made in \cite{Dollar2012Pedestrian}: carefully combining multiple features often improves detection
performance.
The new multiple channel detector outperforms the state-of-the-art by a large margin.
Despite its simplicity, our new approach outperforms
all reported pedestrian detectors, including several
complex detectors such as LatSVM \cite{Felzenszwalb2010Object}
(a part-based approach which models unknown parts as latent variables),
ConvNet \cite{Sermanet2013Pedestrian} (deep hierarchical models)
and DBN-Mut \cite{Ouyang2013Modeling}
(discriminative deep model with mutual visibility relationship).

Doll\'ar \etal propose to compare different detectors using the miss rate
performance at $1$ false positive per image
(FPPI) as a reference point \cite{Dollar2009Pedestrian}.
This performance metric was later revised to the {\em log-average miss rate}
in the range $0.01$ to $1$ FPPI
as this better summarizes
practical
detection performance \cite{Dollar2012Pedestrian}.
This performance metric is also similar to the average precision
reported in text retrieval and PASCAL VOC challenge.
As the performance is assessed over the partial range of
false positives, the performance
of the classifier
outside this range
is ignored as it is not of practical interest.
Many proposed pedestrian detectors optimize the miss rate
over the complete range of false positive
rates, however,
 and can thus produce suboptimal results
both in practice, and in terms of
the log-average miss rate.
In this paper, we address this problem by optimizing the log-average
miss rate performance measure
directly, and
in a more principled manner.
This is significant because it ensures that the detector achieves its best performance within
the range of practical significance, rather than over the whole range
of false positive rates, much of which would be of no practical value.
The approach proposed ensures that the performance is optimized not under
the full ROC curve but only
within the range of practical interest,
thus concentrating performance where it counts, and achieving significantly better results in practice.

\paragraph{Main contributions}
(1) We propose a novel approach to extract low-level visual features
based on spatial pooling for the problem of pedestrian detection.
Spatial pooling has been successfully applied in
sparse coding for generic image classification problems.
The new feature is simple yet outperforms the original covariance
descriptor of \cite{Tuzel2008Pedestrian} and
LBP descriptor of \cite{Wang2009HOG}.
\revised{
(2) We discuss several factors that affect the performance
of boosted decision tree classifiers for pedestrian detection.
Our new design leads to a further improvement in log-average miss rate.
}
(3) Empirical results show that the new approach, which
combines our proposed features with existing features \cite{Dollar2014Fast,Wang2009HOG}
and optimizes the log-average miss rate measure,
outperforms all previously reported pedestrian detection results
and achieves state-of-the-art performance on INRIA, ETH,
TUD-Brussels and Caltech-USA pedestrian detection benchmarks.

\paragraph{Related work}
Numerous pedestrian detectors have been proposed over the past
decade along with newly created pedestrian detection benchmarks such as
INRIA, ETH, TUD-Brussels, Caltech and Daimler Pedestrian data sets.
We refer the reader to \cite{Dollar2012Pedestrian} for an excellent
review on pedestrian detection frameworks and benchmark data sets.
In this section, we briefly discuss several recent state-of-the-art
pedestrian detectors that are not covered in \cite{Dollar2012Pedestrian}.

Sermanet \etal train a pedestrian detector using a
convolutional network model \cite{Sermanet2013Pedestrian}.
Instead of using hand designed features, they propose to
use unsupervised sparse auto encoders to automatically learn features
in a hierarchy.
Experimental results show that their detector achieves competitive results
on major benchmark data sets.
Benenson \etal investigate different low-level aspects of
pedestrian detection \cite{Benenson2013Seeking}.
The authors show that by properly tuning low-level features,
such as
feature selection, pre-processing the raw image
and classifier training,
it is possible to reach state-of-the-art results on major benchmarks.
From their paper, one key observation that significantly improves
the detection performance is to
apply image normalization to the test image before extracting features.

Lim \etal propose novel mid-level features, known as sketch tokens \cite{Lim2013Sketch}.
The feature is obtained from hand drawn sketches in natural images and
captures local edge structure such as straight lines, corners, curves,
parallel lines, \etc
They combine their proposed features with channel features
of \cite{Dollar2009Integral} and train a boosted detector.
By capturing both simple and complex edge structures, their detector
achieves the state-of-the-art result on the INRIA test set.
Park \etal propose new motion features for detecting pedestrians
in a video sequence \cite{Park2013Exploring}.
By factoring out camera motion and combining
their proposed motion features with channel features \cite{Dollar2009Integral},
the new detector achieves a five-fold reduction in false positives
over previous best results on the Caltech pedestrian benchmark.

\section{Our approach}
Despite several
important
work on object detection, the most practical and successful
pedestrian detector is still the sliding-window based method of Viola and Jones \cite{Viola2004Robust}.
Their method consists of two main components: feature extraction and the AdaBoost classifier.
For pedestrian detection, the most commonly used features
are HOG \cite{Dalal2005HOG} and HOG+LBP \cite{Wang2009HOG}.
Doll\'ar \etal propose Aggregated Channel Features (\nACF) which combine
gradient histogram (a variant of HOG), gradients and LUV \cite{Dollar2014Fast}.
ACF uses the same channel features as \ChnFtrs \cite{Dollar2009Integral},
which is shown to outperform HOG \cite{Benenson2013Seeking,Dollar2009Integral}.

To train the classifier, the procedure known as bootstrapping is often applied, which
harvests hard negative examples and re-trains the classifier.
Bootstrapping can be repeated several times.
It is shown in \cite{Walk2010New}  that at least two bootstrapping iterations
are required for the classifier to achieve good performance.
In this paper, we build our detection framework based on \cite{Dollar2014Fast}.
We first propose the new feature type based on a modified low-level descriptor
and spatial pooling.
We then discuss how the miss rate performance measure can be further
improved using structural SVM.
Finally, we discuss our modifications to \cite{Dollar2014Fast} in order to achieve
state-of-the-art detection results on most  benchmark data sets.

\subsection{Spatially pooled features}

\revised{
Spatial pooling has been proven to be invariant to various image transformations and
demonstrate better robustness to noise \cite{Boureau2011Ask,Chatfield2011Devil,Coates2011Importance}.
Several empirical results have indicated that a pooling operation can greatly improve the recognition performance.
Pooling combines several visual descriptors obtained at nearby locations into some statistics that
better summarize the features over some region of interest (pooling region).
The new feature representation preserves visual information over a local neighbourhood
while discarding irrelevant details and noises.
Combining max-pooling with unsupervised feature learning methods have led to
state-of-the art image recognition performance on several object recognition tasks.
Although these feature learning methods have shown promising results over hand-crafted features,
computing these features from learned dictionaries is still a time-consuming process
for many real-time applications.
In this section, we further improve the performance of low-level features by adopting
the pooling operator commonly applied in unsupervised feature learning.
This simple operation can enhance the feature robustness to noise and image transformation.
In the following section, we investigate two visual descriptors which have shown to complement HOG
in pedestrian detection, namely covariance descriptors and LBP.
It is important to point out here that our approach is not limited to these two features,
but can be applied to any low-level visual features.
}

\paragraph{Covariance matrix}
A covariance matrix is positive semi-definite.
 It provides a measure of the relationship between two or more sets of variates.
The diagonal entries of covariance matrices represent the variance of each feature and
the non-diagonal entries represent the correlation between features.
The variance measures the deviation of low-level features from the mean and provides
information related to the distribution of low-level features.
The correlation provides the relationship between multiple low-level features within the region.
In this paper,
we follow the feature representation as proposed in \cite{Tuzel2008Pedestrian}.
However, we introduce an additional edge orientation which considers
the sign of intensity derivatives.
Low-level features used in this paper are:
\begin{align}
  \notag
\left[  x, \; y, \; |I_x|, \; |I_y|, \; |I_{xx}|, \; |I_{yy}|, \; M, \; O_1, \; O_2 \right]
\end{align}
where $x$ and $y$ represent the pixel location, and
$I_x$ and $I_{xx}$ are first and second intensity derivatives along the $x$-axis.
The last three terms are the
gradient
 magnitude
($M = \sqrt{I_x^2 +I_y^2}$),
edge orientation as in \cite{Tuzel2008Pedestrian} ($O_1 = \arctan ( |I_x| / | I_y| )$)
and an additional edge orientation $O_2$ in which,
\begin{align}
  \notag
O_2 =
            \begin{cases}
                \atan2(I_y,I_x)  \quad& \text{if} \; \atan2(I_y,I_x) > 0, \\
                \atan2(I_y,I_x) + \pi  \quad& \text{otherwise.}
            \end{cases}
\end{align}
The orientation $O_2$ is mapped over the interval $[0,\pi]$.
Although some $O_1$ features might be redundant after introducing $O_2$,
these features would not deteriorate the performance as they
will not be selected by the weak learner.
Our preliminary experiments show that using $O_1$ alone yields
slightly worse performance than combining $O_1$ and $O_2$.
With the defined mapping, the input image is mapped to a $9$-dimensional feature image.
The covariance descriptor of a region is a $9 \times 9$ matrix, and due to symmetry,
only the upper triangular part is stored,
which has only $45$ different values.

\paragraph{LBP}
\revised{
Local Binary Pattern (LBP) is a texture descriptor that represents the binary code of each image patch
into a feature histogram \cite{Ojala2002Multi}.
The standard version of LBP is formed by thresholding the $3 \times 3$-neighbourhood of each pixel with the centre pixel's value.
All binary results are combined to form an $8$-bit binary value ($2^8$ different labels).
The histogram of these $256$ different labels can be used as texture descriptor.
The LBP descriptor has shown to achieve good performance in many texture classification \cite{Ojala2002Multi}.
In this work, we transform the input image from the RGB color space
to LUV space and apply LBP to the luminance (L) channel.
We adopt an extension of LBP, known as the uniform LBP, which can better filter out noises \cite{Wang2009HOG}.
The uniform LBP is defined as the binary pattern that contains at most two bitwise transitions from $0$ to $1$ or vice versa.
}

\begin{figure}[t]
    \centering
        \includegraphics[width=0.9\textwidth,clip]{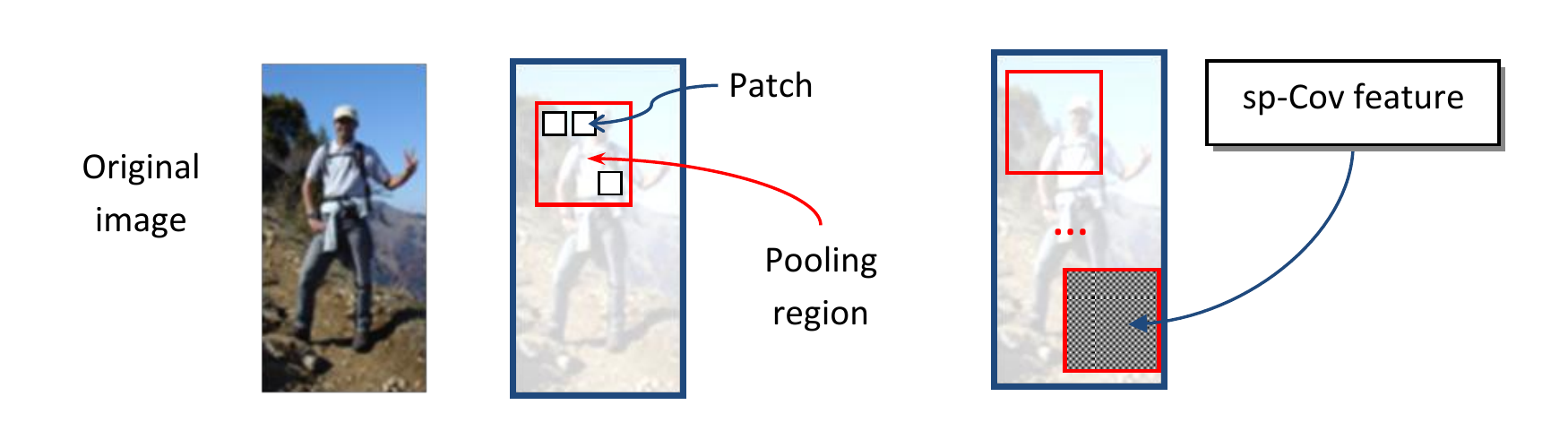}
    \caption{
    Architecture of our pooled features.
    In this example, \SCOV are extracted from each fixed sized pooling region.
    }
    \label{fig:illus}
\end{figure}

\paragraph{Spatially pooled covariance}
In this section, we improve the spatial invariance and robustness of the
original covariance descriptor by applying the operator known as spatial pooling.
There exist two common pooling strategies in the literature:
average pooling and max-pooling.
We use max-pooling as it has been shown to outperform average
pooling in image classification \cite{Coates2011Importance,Boureau2011Ask}.
We divide the image window into {\em dense patches}
(refer to Fig.~\ref{fig:illus}).
For each patch, covariance features are calculated over pixels within
the patch.
For better invariance to translation and deformation,
we perform spatial pooling over a fixed-size spatial region
({\em pooling region}) and use the obtained results
to represent covariance features
in the pooling region.
The pooling operator thus summarizes multiple covariance matrices
within each pooling region into a
single matrix which represents covariance information.
We refer to the feature extracted from each pooling region
as spatially pooled covariance (\nSCOV) feature.
Note that extracting covariance features in each patch
can be computed efficiently using the integral image trick \cite{Tuzel2006Region}.
Our \SCOV
differs from covariance features in
\cite{Tuzel2008Pedestrian} in the following aspects:

1. We apply spatial pooling to a set of covariance descriptors in the pooling region.
  To achieve this, we ignore the geometry of covariance matrix
  and stack the upper triangular part of the covariance matrix into a vector
  such that pooling
	is carried out
	on the vector space.
  For simplicity,
	we carry out pooling over
        a square image region of fixed resolution.
  Considering pooling over a set of arbitrary rectangular regions as in \cite{Jia2012Beyond}
	is likely to
	further improve the performance of our features.

2. Instead of normalizing the covariance descriptor of each patch based on the whole
  detection window \cite{Tuzel2008Pedestrian},
  we calculate the correlation coefficient within each patch.
  The correlation coefficient returns the value in the range $[-1,1]$.
  As each patch is now independent, the feature extraction can be done
  in parallel on the GPU.

\paragraph{Implementation}
We extract \SCOV using multi-scale patches with the following sizes:
$8 \times 8$, $16 \times 16$ and $32 \times 32$ pixels.
Each scale will generate a different set of visual descriptors.
Multi-scale patches have also been used in \cite{Bo2013Multipath}.
In this paper, the use of multi-scale patches is important as it expands
the richness of our feature representations and enables us to capture
human body parts at different scales.
In our experiments, we set the patch spacing stride (step-size)
to be $1$ pixel.
The pooling region is set to be  $4 \times 4$ pixels
and the pooling spacing stride is set to $4$ pixels in our experiments.

\paragraph{Spatially pooled LBP}
\revised{
Similar to \SCOV, we divide the image window into small patches and extract LBP over pixels within the patch.
The histogram, which represents the frequency of each pattern occurring, is computed over the patch.
For better invariance to translation, we perform spatial pooling over a pooling region and use the obtained results
to represent the LBP histogram in the pooling region.
We refer to the new feature as spatially pooled LBP (\nSLBP) feature.
}

\paragraph{Implementation}
\revised{
We apply the LBP operator on the $3 \times 3$-neighbourhood at each pixel.
The LBP histogram is extracted from a patch size of $4 \times 4$ pixels.
We extract the 58-dimension LBP histogram using a C-MEX implementation of \cite{Vedaldi2010VLFeat}.
For \SLBP, the patch spacing stride, the pooling region and the pooling spacing stride are
set to $1$ pixel, $8 \times 8$ pixels and $4$ pixels, respectively.
We also experiment with combining the LPB histogram extracted from multi-scale patches
but only observe a slight improvement in detection performance at a much higher feature extraction time.
Instead of extracting LBP histograms from multi-scale patches,
we combine \SLBP and LBP as channel features.
}

\paragraph{Discussion}
Although we make use of spatial pooling, our approach
differs significantly from the unsupervised feature learning pipeline,
which has been successfully
applied to image classification problem \cite{Bo2013Multipath,Yang2009Linear}.
Instead of pooling encoded features over a pre-trained dictionary,
we compute \SCOV and \SLBP by performing pooling directly on
covariance and LBP features extracted from local patches.
In other words, our proposed approach removes the dictionary learning and
feature encoding from the conventional unsupervised feature learning
\cite{Bo2013Multipath,Yang2009Linear}.
The advantage of our approach over conventional feature learning
is that our features have much less dimensions than
the size of visual words often used in generic image classification \cite{Yang2009Linear}.
Using too few visual words can significantly degrade
the recognition performance as reported in \cite{Chatfield2011Devil} and using too many visual words would
lead to very high-dimensional features and thus make the classifier training become computationally infeasible.

\subsection{Optimizing the partial area under ROC curve}
\revised{
As the performance of the detector is usually measured using
the log-average miss rate, we optimize the \pAUC (the partial AUC)
between any two given false positive rates $[\alpha,\beta]$,
similar to the work of \cite{Paul2013Efficient}.
Unlike \cite{Paul2013Efficient}, in which weak learners are
selected based on the \pAUC criterion,
we use AdaBoost to select weak learners as it is more efficient.
In order to achieve the best performance, we build a feature vector
from the weak learners' output and learn the \pAUC classifier in
the final stage.
For each predicted positive patch, the
confidence score is re-calibrated based on this \pAUC classifier.
This post-learning step is similar to the work of \cite{Wu2008Fast},
in which the authors learn the asymmetric classifier from the output of AdaBoost's weak learners
to handle the node learning goal in the cascade framework.
}

The \pAUC risk for a scoring function $f(\cdot)$ between two pre-specified
\FPR $[\alpha,\beta]$ can be defined \cite{Narasimhan2013SVM} as :
\begin{align}
    \label{EQ:AUC}
        \hat{R}_{\zeta}(f) = {\textstyle \sum}_{i=1}^m
                {\textstyle \sum}_{j=\jalpha+1}^{\jbeta} \v1 ( f(\vxip) < f(\vxjfzm) ).
\end{align}
Here $\vxip$ denotes the $i$-th positive training instance
and
$\vxjfzm$ denotes the $j$-th negative training instance sorted by $f$ in
the set $\zeta \in \sZ_{\beta}$.
Both $\vxip$ and $\vxjfzm$ represent the output vector
of weak classifiers learned from
AdaBoost.
Clearly \eqref{EQ:AUC} is minimal when all positive samples, $\{ \vxip \}_{i=1}^m$,
are ranked above
$\{ \vxjfzm \}_{j=\jalpha+1}^{\jbeta}$,
which represent negative samples
in our prescribed false positive range $[\alpha,\beta]$
(in this case, the log-average miss rate would be zero).
The structural SVM framework can be adopted to optimize the \pAUC risk by
considering a classification problem of all $m \times \jbeta$ pairs of
positive and negative samples.
In our experiments, the \pAUC classifier is trained once at the final bootstrapping iteration
and most of the computation time is spent in extracting features
and bootstrapping hard negative samples.
See the supplementary material for more details on the structural SVM problem.

\subsection{Design space}
\label{sec:design_space}
In this section, we further investigate the experimental
design of the \ACF detector \cite{Dollar2014Fast}.
For experiments on shrinkage and spatial pooling, we use the proposed
\SCOV as channel features.
For experiments on the depth of decision trees, we use channel features
of \cite{Dollar2014Fast}.
All experiments are carried out using AdaBoost with the shrinkage
parameter of $0.1$ as a strong classifier and
level-$3$ decision trees as weak classifiers (if not specified otherwise).
We use three bootstrapping stages and the final
model consists of $2048$ weak classifiers with soft cascade.
We heuristically set the soft cascade's reject threshold to be $-10$ at every node.
We trained all detectors using the INRIA training set and evaluated the detector on
INRIA, ETH and TUD-Brussels benchmark data sets.

\paragraph{Shrinkage}
Hastie \etal show that the accuracy of boosting can be further improved by
applying a weighting coefficient known as shrinkage \cite{Hastie2009Elements}.
The explanation given
in \cite{Friedman2000Additive} is that a shrinkage version of boosting
simply converges to the $\ell_1$ regularized solution.
It can also be viewed as another form of regularization for boosting.
At each
iteration, the weak learner's coefficient is updated by
\begin{align}
  F_t(\vx) = F_{t-1}(\vx) + \nu \cdot {\omega}_t h_t(\vx)
\end{align}
Here $h_t(\vx)$ is AdaBoost's weak learner at the $t$-th iteration
and $\omega_t$ is the weak learner's
coefficient at the $t$-th iteration.
$\nu \in (0, 1]$ can be viewed as a learning rate parameter.
The smaller the value of $\nu$, the higher the overall accuracy
as long as the number of iterations is large enough.
\cite{Friedman2000Additive} report that shrinkage often produces
 better generalization performance compared to linear search algorithms.
We compare four different shrinkage parameters from $\{0.05$, $0.1$, $0.2$, $0.5\}$
with the conventional AdaBoost.
When applying shrinkage, we lower the soft cascade's reject threshold by
a factor of $\nu$ as weak learners' coefficients have been diminished by
a factor of $\nu$.
The log-average miss rate of different detectors is shown in Table~\ref{tab:shrinkage}.
We observe that applying a small amount of shrinkage ($\nu \leq 0.2$) often
improves the detection performance.
From Table~\ref{tab:shrinkage}, setting the shrinkage value to be too small ($\nu = 0.05$)
without increasing the number of weak classifiers
can hurt the performance as the number of boosting iterations is not large enough
for the boosting to converge.
For the rest of our experiments, we set the shrinkage parameter to be $0.1$
as it gives a better trade-off between the performance and the number of weak classifiers.

\begin{table}[t]
  \centering
  \caption{
  Log-average miss rate when varying shrinkage parameters.
  Shrinkage can further improve the final detection performance.
  $\dag$ The model consists of $4096$ weak classifiers
  while other models consist of $2048$ weak classifiers
  }
  \scalebox{1}
  {
  \begin{tabular}{l|ccc|c}
  \hline
   Shrinkage & INRIA & ETH & TUD-Br. & Avg. \\
  \hline
  \hline
   None & $14.4\%$ & $40.8\%$ & $48.7\%$ & $34.6\%$ \\
   $\nu = 0.5$ & $12.5\%$ & $43.7\%$ & $50.3\%$ & $35.5\%$ \\
   $\nu = 0.2$ & $11.6\%$ & $41.4\%$ & $50.4\%$ & $34.4\%$\\
   $\nu = 0.1$ & $12.8\%$ & $42.0\%$ & $47.8\%$ & $\mathbf{34.2\%}$\\
   $\nu = 0.05$ & $14.0\%$ & $43.1\%$ & $51.4\%$ & $36.2\%$\\
   $\nu = 0.05^{\dag}$ & $12.8\%$ & $42.6\%$ & $48.6\%$ & $34.7\%$\\
  \hline
  \end{tabular}
  }
  \label{tab:shrinkage}
\end{table}

\paragraph{Spatial pooling}
In this section, we compare the performance of the proposed feature
with
and without
spatial pooling.
For \SCOV and \SLBP without pooling, we extract both low-level visual features
with the patch spacing stride of $4$ pixels and no pooling is performed.
Using these low-level features and LUV colour features,
we trained four detectors using the INRIA training set.
Log-average miss rates of both features are shown in Table~\ref{tab:pooling}.
We observe that it is beneficial to apply spatial pooling as
it increases the robustness of the features against
small deformations and translations.
We observe a reduction in miss rate by more than one percent on the INRIA test set.
Since we did not combine \SLBP with HOG as in \cite{Wang2009HOG},
\SLBP performs slightly worse than \SCOV.

\begin{table}[bt]
  \caption{
  Log-average miss rate of our features with and without
  applying spatial pooling.
  }
  \centering
  \scalebox{1}
  {
  \begin{tabular}{l|ccc|ccc}
  \hline
    & \multicolumn{3}{c|}{Covariance} &  \multicolumn{3}{c}{LBP} \\
  \hline
     & INRIA & ETH & TUD-Brussels & INRIA & ETH & TUD-Brussels \\
  \hline
  \hline
   without pooling & $14.2\%$ & $42.7\%$ & $48.6\%$ &  $25.8\%$ & $47.8\%$ & $\mathbf{55.5\%}$ \\
   with pooling & $\mathbf{12.8\%}$ & $\mathbf{42.0\%}$ & $\mathbf{47.8\%}$ &  $\mathbf{23.7\%}$ & $\mathbf{46.2\%}$ & $55.8\%$\\
  \hline
  \end{tabular}
  }
  \label{tab:pooling}
\end{table}

\begin{figure}[t]
    \centering
        \includegraphics[width=0.325\textwidth,clip]{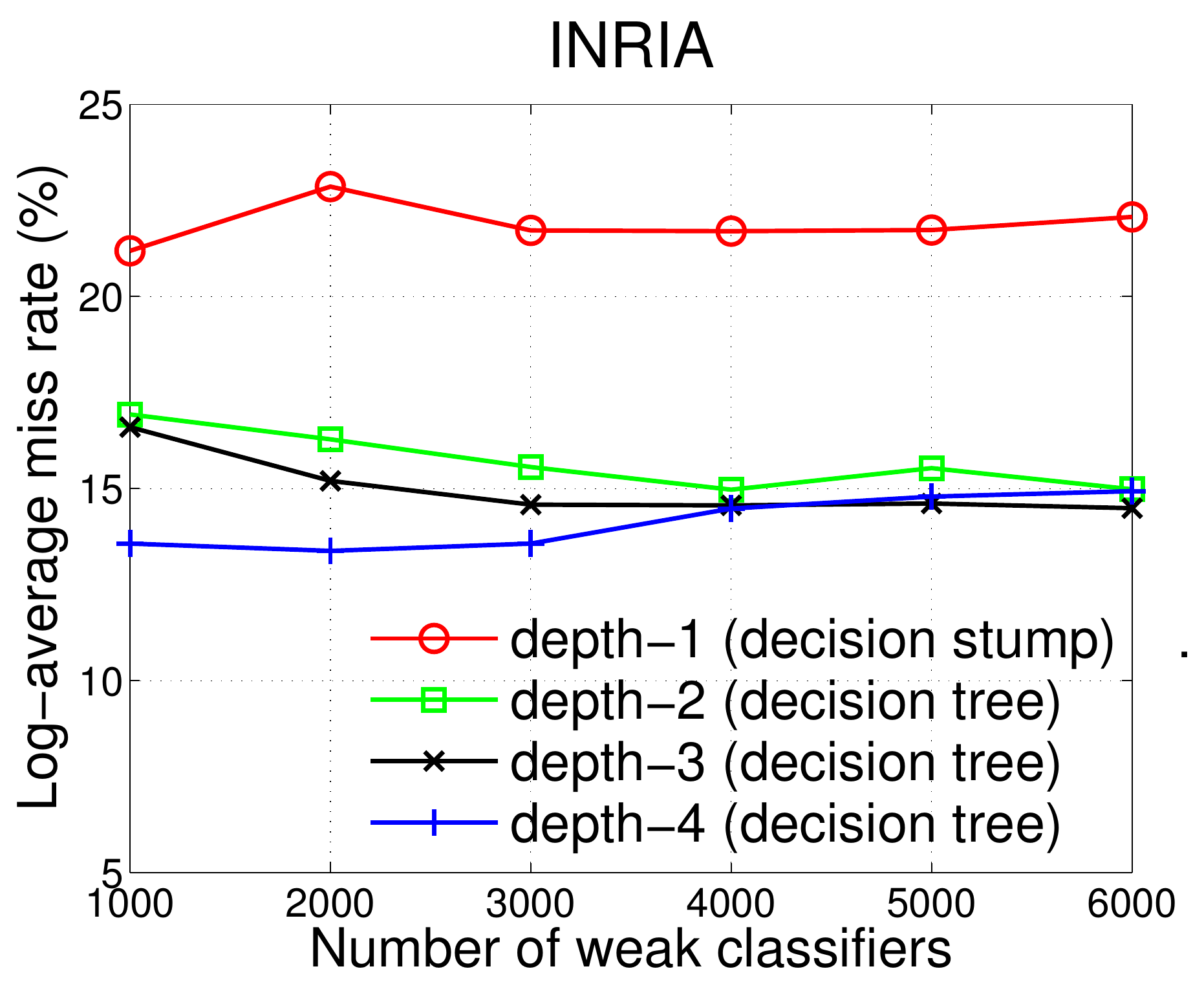}
        \includegraphics[width=0.3\textwidth,clip]{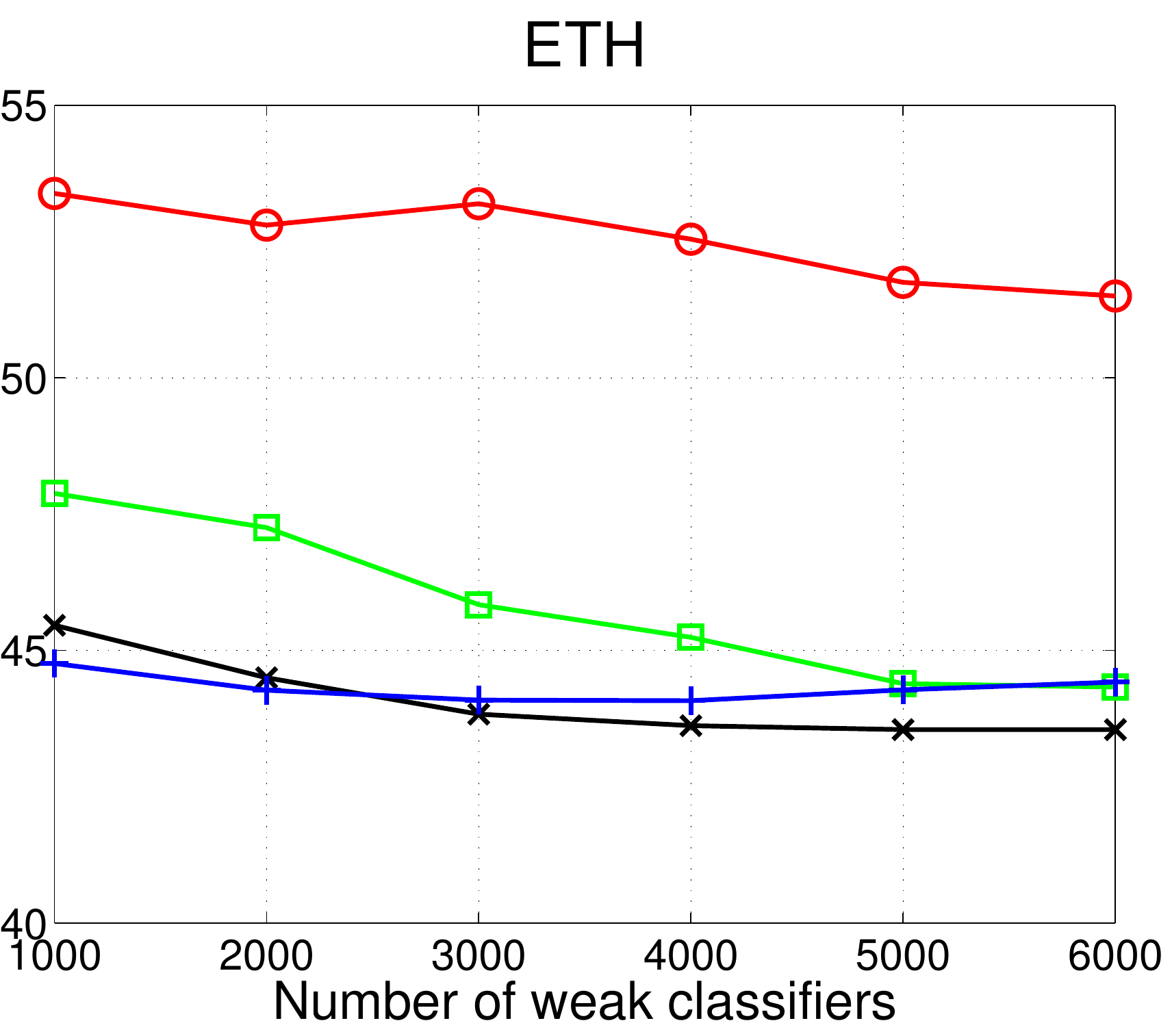}
    \caption{
    Log-average miss rates of different tree depths on INRIA (left) and ETH (right) benchmark data sets.
    }
    \label{fig:weaklearner}
\end{figure}

\paragraph{Depth of decision trees}
Both \cite{Dollar2009Integral} and \cite{Benenson2013Seeking} report that the depth-2 decision tree
produces the best performance in their experiments.
However, we observe that the depth-3 decision tree offers better generalization performance.
To conduct our experiments, we trained $4$ different pedestrian detectors with decision trees
of depth $1$ (decision stump) to $4$ (containing $15$ stumps).
Our experiments are based on the \ACF detector of \cite{Dollar2014Fast} which combines
gradient histogram (O), gradient (M) and LUV features.
The \ACF detector linearly quantizes feature values into $256$ bins
to speed up the conventional decision tree training \cite{Appel2013Quickly}.
We trained the pedestrian detector using the INRIA training set and evaluated the detector on both
INRIA and ETH benchmark data sets.
Fig.~\ref{fig:weaklearner}
plots the log-average miss rate on the vertical axis and the number of weak classifiers on
the horizontal axis.
We observe that the
pedestrian detection performance improves as we increase the depth of decision trees.
Similar to \cite{Benenson2013Seeking}, we observe that using decision stumps as weak learners
can lead to significant underfitting, \ie, the weak learner can not separate pedestrian patches
from non-pedestrian patches.
On the other hands, setting the tree depth to be larger than two can lead to a
performance improvement, especially on the ETH data set.
For the rest of our experiments, we set the depth of decision trees to be three as it achieves
good generalization performance and is faster to train than the depth-$4$ decision tree.

\section{Experiments}

We train two detectors: one using the INRIA training set and one using the Caltech-USA training set.
For INRIA, each pedestrian training sample is scaled to a resolution of $64 \times 128$ pixels.
Negative patches are collected from INRIA background images.
We follow the work of \cite{Dollar2014Fast}
to train the boosted pedestrian detector.
Each detector is trained using three bootstrapping stages
and consists of $2048$ weak classifiers.
The detector trained on the INRIA training set is evaluated on
all benchmark data sets except the Caltech-USA
test set\footnote{\cite{Park2013Exploring} report a performance
improvement on the Caltech-USA when they retrain the detector
using the Caltech-USA training set.
We follow the setup discussed in \cite{Park2013Exploring}.
}.
On both ETH and TUD-Brussels data sets, we apply the automatic colour
equalization algorithms (ACE) \cite{Rizzi2003New}
before we extract channel features \cite{Benenson2013Seeking}.
We upscale both ETH and TUD-Brussels test images to $1280 \times 960$ pixels.
For Caltech-USA, the resolution of the pedestrian model is set to
$32 \times 64$ pixels.
We exclude occluded pedestrians from the Caltech training set \cite{Park2013Exploring}.
Negative patches are collected from the Caltech-USA training set with
pedestrians  cropped out.
To obtain final detection results,
greedy non-maxima
suppression is applied
with the default parameter as in Addendum of \cite{Dollar2009Integral}.
We use the log-average miss rate to summarize the detection performance.
For the rest of our experiments, we evaluate our pedestrian detectors on the
reasonable subset (pedestrians are at least $50$ pixels in height
and at least $65\%$ visible).

\subsection{Improved covariance descriptor}
In this experiment, we evaluate the performance of the proposed \SCOV.
\SCOV consists of $9$ low-level image statistics.
We exclude the mean and variance of two image statistics (pixel locations at x and y co-ordinates)
since they do not capture discriminative information.
We also exclude the correlation coefficient between pixel locations at x and y co-ordinates.
Hence there is a total of $136$ channels ($7$ low-level image statistics +
$3 \cdot 7$ variances + $3 \cdot 35$ correlation coefficients + $3$ LUV color
channels)\footnote{Note here that we extract
covariance features at $3$ different scales.}.
It is important to note here that our features and weak classifiers
are different from
those
in \cite{Tuzel2008Pedestrian}.
\cite{Tuzel2008Pedestrian} calculates the covariance distance in the Riemannian manifold.
As eigen-decomposition is performed, the approach of \cite{Tuzel2008Pedestrian} is computationally expensive.
We speed up the weak learner training by proposing our modified covariance features and
train the weak learner using the decision tree.
The new weak learner is not only simpler than \cite{Tuzel2008Pedestrian} but also highly effective.

\begin{figure}[t]
    \centering
        \includegraphics[width=0.45\textwidth,clip]{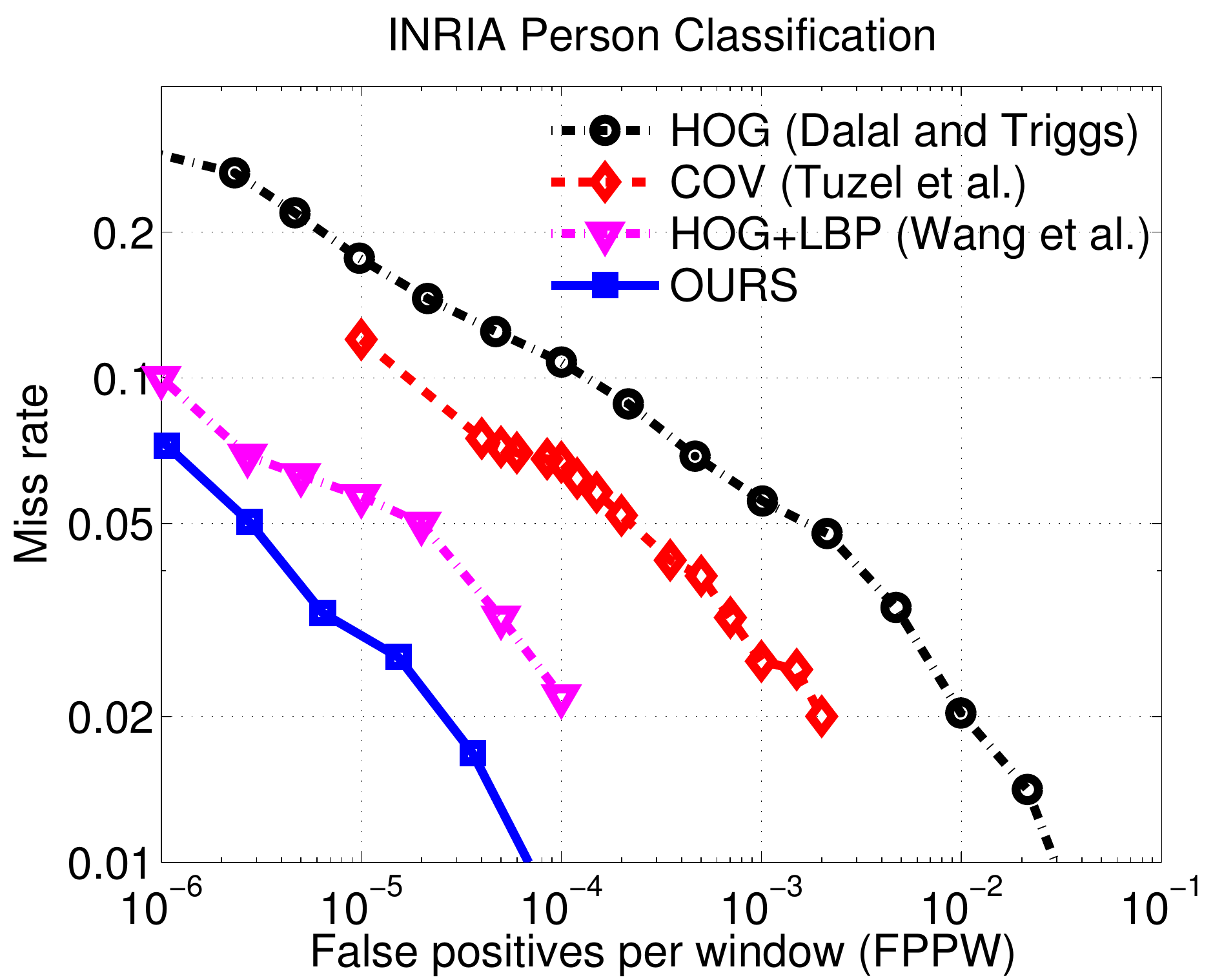}
    \caption{
    ROC curves of our \SCOV features and
    the conventional covariance detector \cite{Tuzel2008Pedestrian}
    on the INRIA test image.
    }
    \label{fig:cov_fppw}
\end{figure}

We compare our detector with the original covariance descriptor \cite{Tuzel2008Pedestrian} in Fig.~\ref{fig:cov_fppw}.
We plot HOG \cite{Dalal2005HOG} and HOG+LBP \cite{Wang2009HOG} as the baseline.
Similar to the result reported in \cite{Benenson2013Seeking}, where the authors show that
\HOGBOOST reduces the average miss-rate over \HOGSVM by more than $30\%$,
we observe that applying our \SCOV features as the channel features
significantly improves the detection performance over the original covariance detector
(a reduction of more than $5\%$ miss rate at $10^{-4}$ false positives per window).
More experiments on \SCOV with different subset of low-level features,
multi-scale patches and spatial pooling parameters can be found in the supplementary.

Next we compare the proposed \SCOV with \ACF features (M+O+LUV) \cite{Dollar2014Fast}.
Since \ACF uses
fewer
channels than \SCOV,
for a fair comparison,
we increase \nACF's discriminative power by combining
\ACF features with LBP\footnote{In our implementation, we use an extension of
LBP, known as the uniform LBP, which can better filter out noises \cite{Wang2009HOG}.
Each LBP bin corresponds to each channel.} (M+O+LUV+LBP).
The results are reported in Table~\ref{tab:featcomb}.
We observe that \SCOV yields competitive results to M+O+LUV+LBP.
From the table, \SCOV performs better on the INRIA test set, worse on the ETH test set and on par with
M+O+LUV+LBP on the TUD-Brussels test set.
We observe that the best performance is achieved by combining \SCOV and \SLBP with
M+O+LUV.

\begin{table}[tb]
  \caption{
  Log-average miss rates of various feature combinations
  }
  \centering
  \scalebox{1}
  {
  \begin{tabular}{l|c|ccc}
  \hline
     & \# channels & INRIA & ETH & TUD-Br. \\
  \hline
  \hline
    M+O+LUV+LBP        & 68  & $14.5\%$ & $39.9\%$ & $47.0\%$ \\
    \nSCOV+LUV         & 136 &$12.8\%$ & $42.0\%$ & $47.8\%$ \\
    \nSCOV+M+O+LUV     & 143 & $\mathbf{11.2\%}$ & $39.4\%$ & $46.7\%$ \\
    \nSCOV+\nSLBP+M+O+LUV  & 259 & $\mathbf{11.2\%}$ & $\mathbf{38.0\%}$ & $\mathbf{42.5\%}$ \\
  \hline
  \end{tabular}
  }
  \label{tab:featcomb}
\end{table}

\begin{figure}[t]
    \centering
        \includegraphics[width=0.45\textwidth,clip]{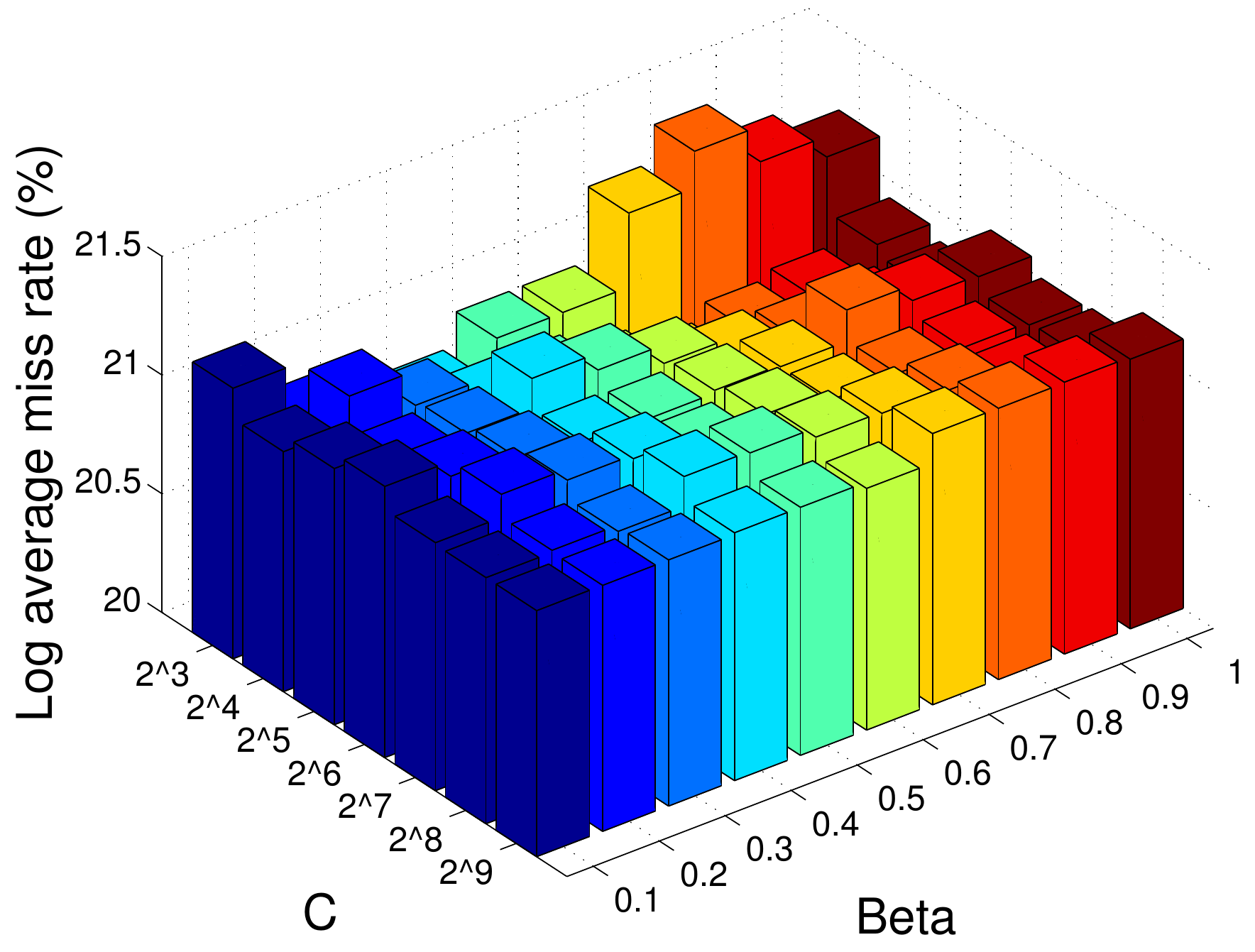}
    \caption{
    Cross-validation results (log-average miss rate) as the \paucstruct regularization parameter $C$ and
    the false positive rate $\beta$ change.
    Without \paucstruct, the detector achieves the miss rate of $21.3\%$.
    The detector with post-tuning ($C = 2^4$ and $\beta = 0.7$) performs best
    with a miss rate of $20.7\%$ (an improvement of $0.6\%$).
    }
    \label{fig:pauc1}
\end{figure}

\begin{figure*}[t]
    \centering
        \includegraphics[width=0.32\textwidth,clip]{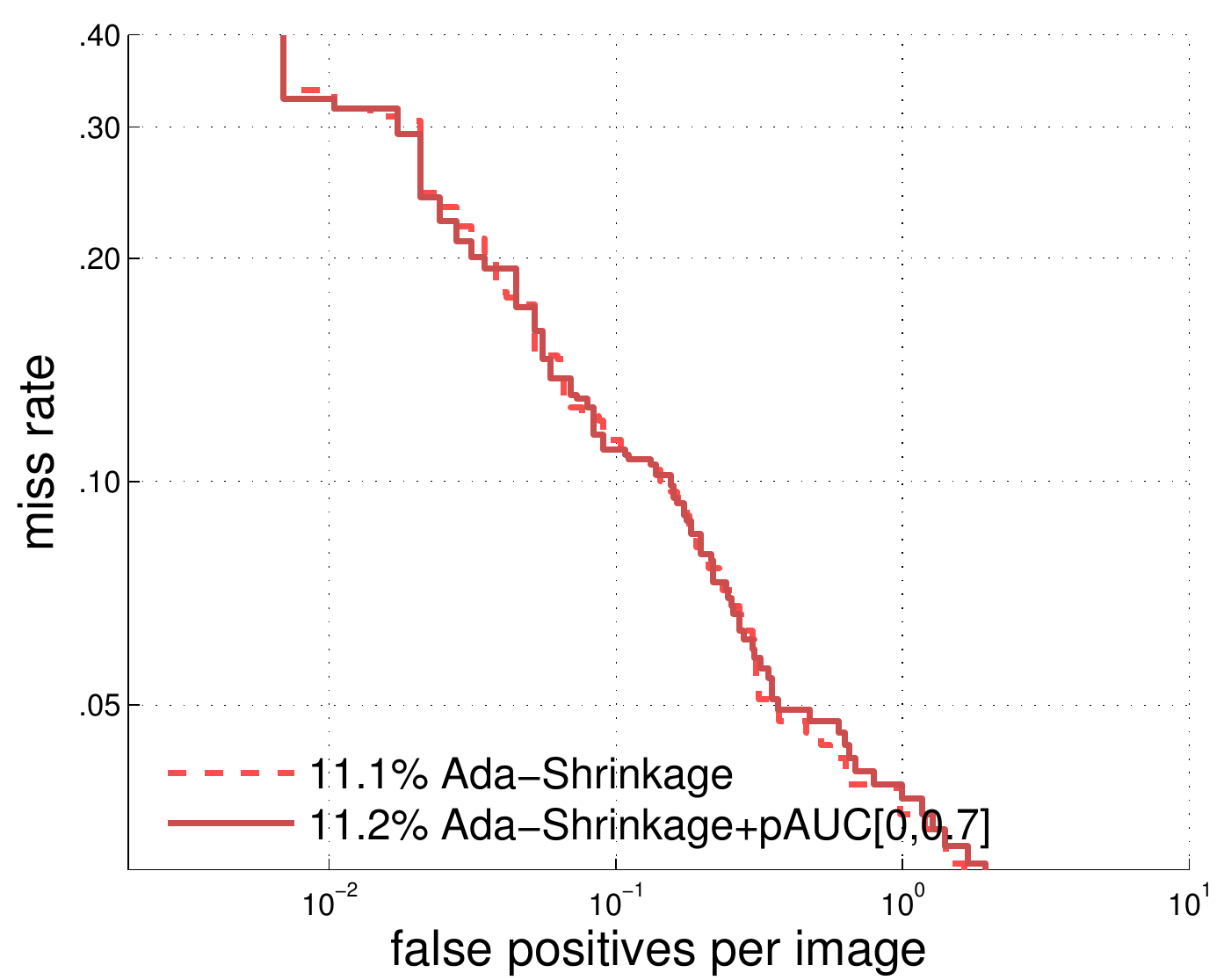}
        \includegraphics[width=0.32\textwidth,clip]{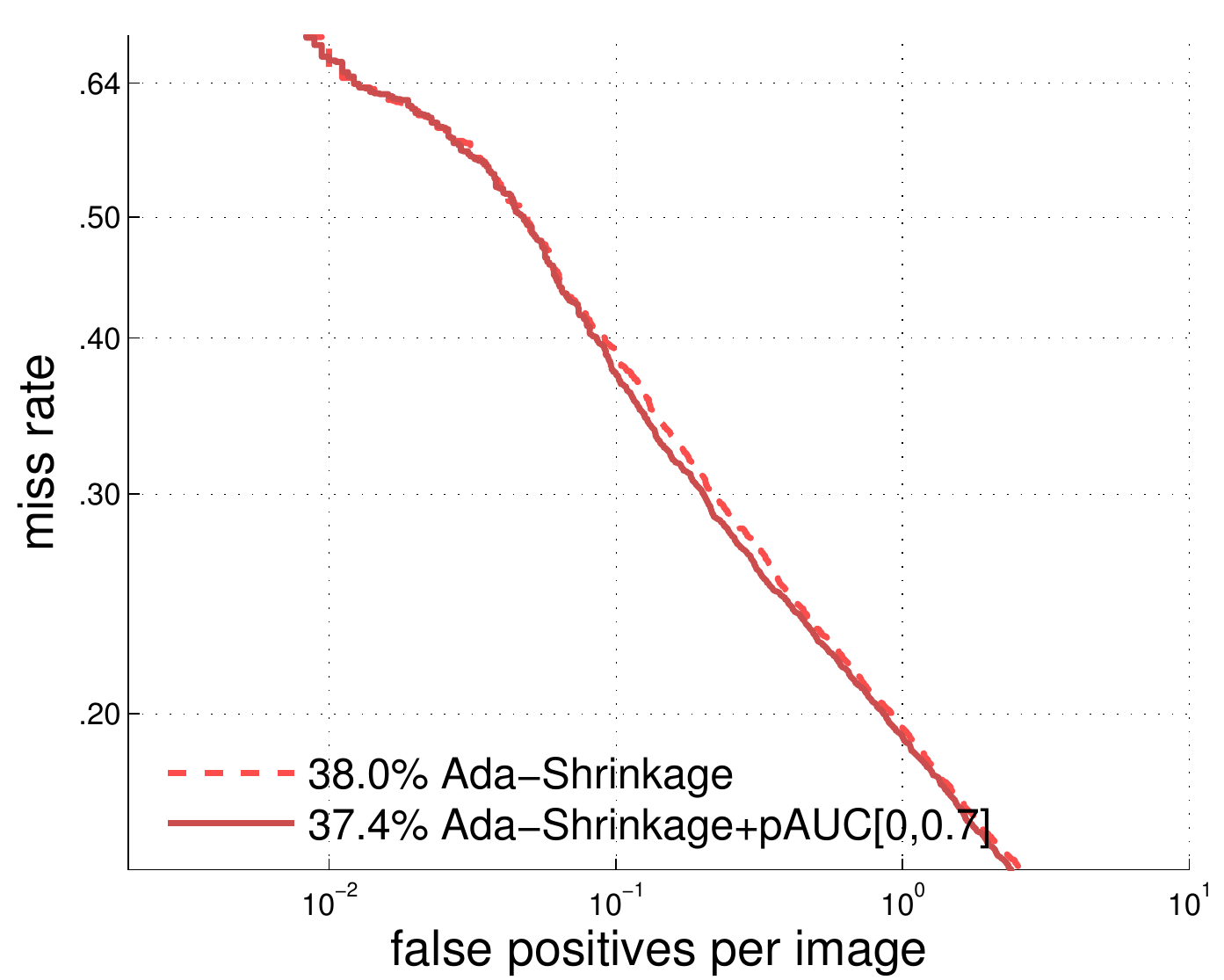}
        \includegraphics[width=0.32\textwidth,clip]{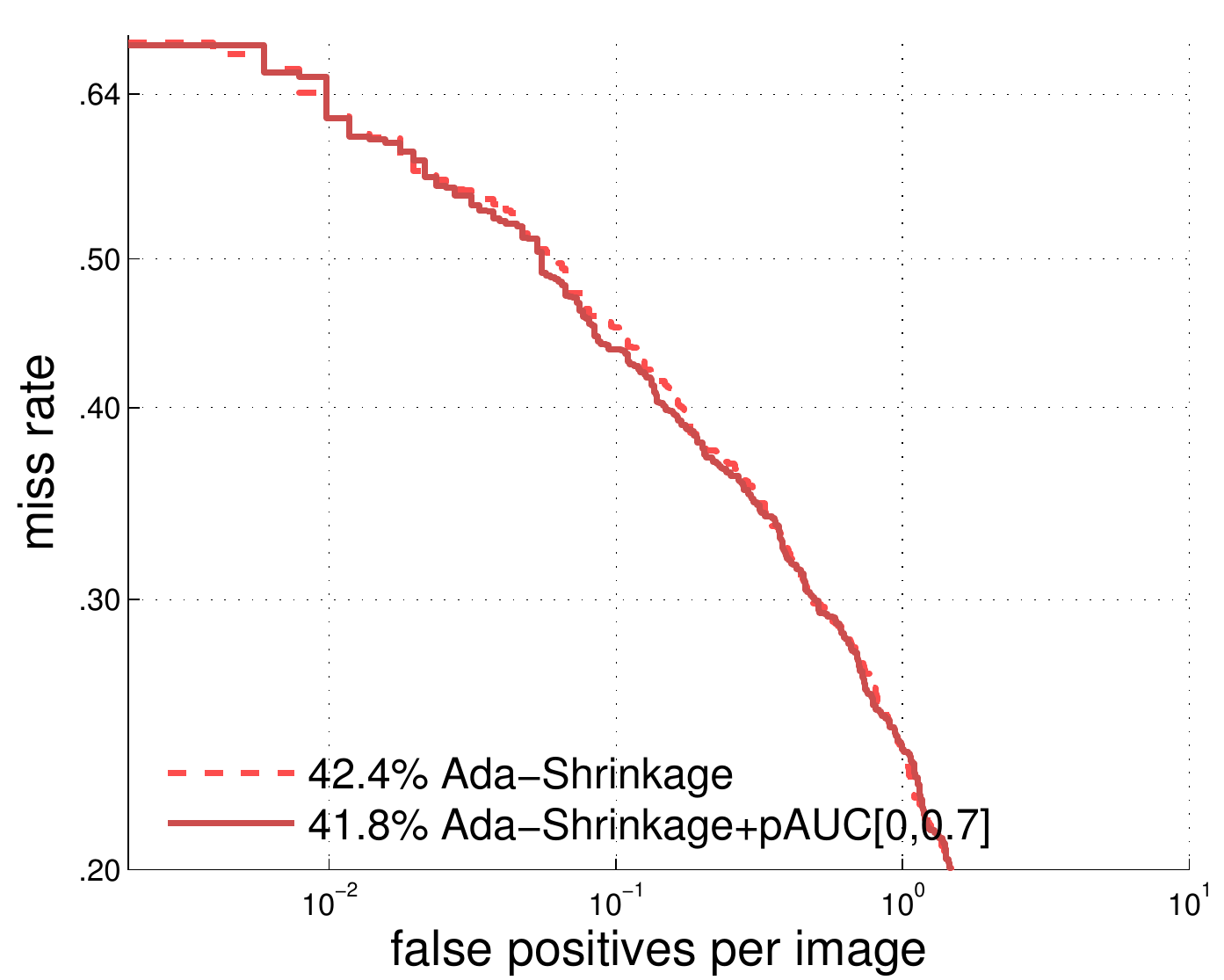}
    \caption{
    Detection performance of our detectors with \paucstruct post-tuning on INRIA ({\em left}),
    ETH ({\em middle}) and TUD-Brussels ({\em right}) benchmark data sets.
    }
    \label{fig:pauc2}
\end{figure*}

\subsection{Improving average miss rate with \paucstruct}
In this experiment, we evaluate the effect of re-calibrating
the final confidence score with \paucstruct.
Instead of using the weighted responses from AdaBoost,
we re-rank the confidence score of predicted pedestrian patches
using a scoring function of \paucstruct.
The performance is calculated by varying the threshold value in the
false positive range of $[0.01, 1]$ FPPI.
Since the partial area under the ROC curve is determine on a
logarithmic scale \cite{Dollar2012Pedestrian}, it is non-trivial
to determine the best \paucstruct parameters $\alpha$ and $\beta$ which
maximize the detection rate between $0.01$ and $1$ false positive per image.
In our experiment, we heuristically set $\alpha$ to be $0$ and perform a
cross-validation to find the best
\paucstruct regularization parameter $C$  (see Supplementary) and
the false positive rate $\beta$.
In this section, we first train the baseline pedestrian detector as discussed
in Section~\ref{sec:design_space}.
The baseline detector achieves the log-average miss rate of $21.3\%$.
Next we perform the post-learning step by re-ranking
the confidence score of positive
and negative samples based on the pAUC criterion.
Using cross-validation on the INRIA training set, the post-learning step
improves the log-average miss rate by $0.6\%$.
Fig.~\ref{fig:pauc1} plots the log average miss rate with respect to
the \paucstruct regularization parameter $C$ and
the false positive rate $\beta$.
From the figure, the following parameters ($C = 2^4$ and $\beta = 0.7$) perform best
with a miss rate of $20.7\%$.
In the next experiment, we evaluate
the detector with the post-learning step on INRIA, ETH and TUD-Brussels benchmark data sets.
ROC curves along with their log-average miss rates between $[0.01, 1]$ FPPI are
shown in Fig.~\ref{fig:pauc2}.
Based on our results, applying \paucstruct improves the log-average miss rate
of the original detector on both ETH and TUD-Brussels benchmark data sets
by $0.6\%$.
However we do not observe an improvement on the INRIA test set.
Our conjecture is that the INRIA test set consists of high-resolution human
in a standing position which might be easier to detect than
those appeared in ETH and TUD-Brussels data sets.
No improvement in detection performance is observed on the INRIA test set
as compared to the detection results on ETH and TUD-Brussels data sets.

\begin{table*}[t]
 \caption{
  Log-average miss rates of various algorithms on INRIA, ETH, TUD-Brussels
  and Caltech-USA test sets.
  The best detector is shown in boldface.
  We train two detectors: one using INRIA training set (evaluated on
  INRIA, ETH and TUD-Brussels test sets) and another one using Caltech-USA
  training set (evaluated on Caltech-USA test set).
  The log-average miss rate of our detection results
  are calculated using the Caltech pedestrian detection benchmark version $3.2.0$.
  $\dag$ Results reported here are taken from
  \url{http://www.vision.caltech.edu/Image_Datasets/CaltechPedestrians/}
  and are slightly different from the one reported in the original paper
  }
  \centering
  \scalebox{0.8}
  {
  \begin{tabular}{l|cccc}
  \hline
   Approach & INRIA & ETH & TUD-Brussels & Caltech-USA\\
  \hline
  \hline
   Sketch tokens \cite{Lim2013Sketch} (Prev. best on INRIA$^{\dag}$)  & $13.3\%$ & N/A & N/A & N/A \\
   DBN-Mut \cite{Ouyang2013Modeling} (Prev. best on ETH$^{\dag}$)  & N/A & $41.1\%$ & N/A & $48.2\%$ \\
   MultiFtr+Motion+2Ped \cite{Ouyang2013Single} (Prev. best on TUD-Brussels) & N/A & N/A & $50.5\%$ & N/A \\
   SDtSVM \cite{Park2013Exploring} (Prev. best on Caltech-USA) & N/A & N/A & N/A & $36.0\%$ \\
   Roerei \cite{Benenson2013Seeking} ($2$-nd best on INRIA$^{\dag}$ \& ETH$^{\dag}$)  & $13.5\%$ & $43.5\%$ & $64.0\%$ & $48.4\%$ \\
  \hline
   Ours (\nSCOV+\nSLBP+M+O+LUV) & $\mathbf{11.1\%}$ & $38.0\%$ & $42.4\%$ & $29.4\%$ \\
   Ours (\nSCOV+\nSLBP+M+O+LUV + \paucstruct) & $11.2\%$ & $\mathbf{37.4\%}$ & $\mathbf{41.8\%}$ & $\mathbf{29.2\%}$ \\
  \hline
  \end{tabular}
  }
  \label{tab:tab1}
\end{table*}

\begin{figure*}
    \centering
        \includegraphics[width=0.46\textwidth,clip]{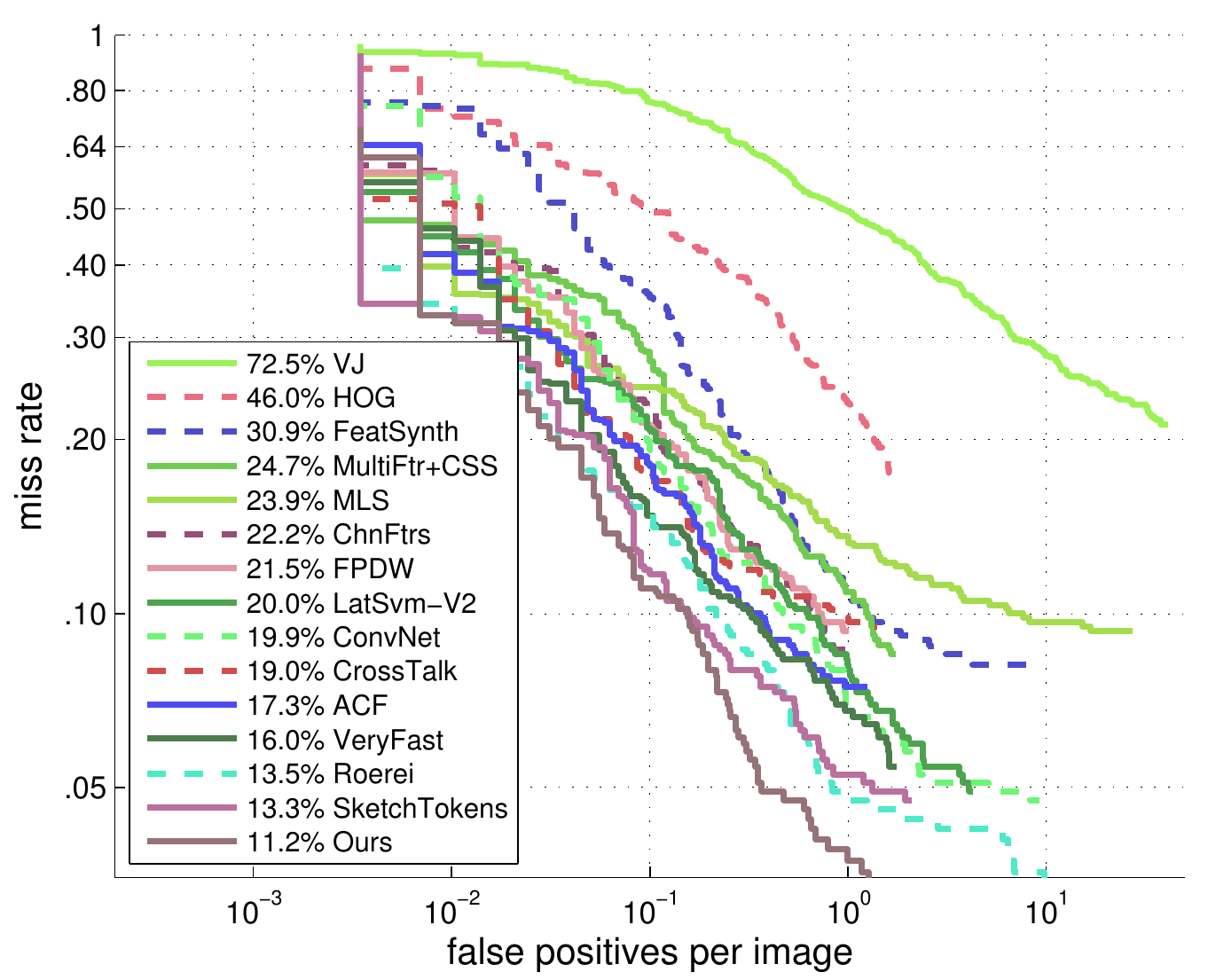}
        \includegraphics[width=0.46\textwidth,clip]{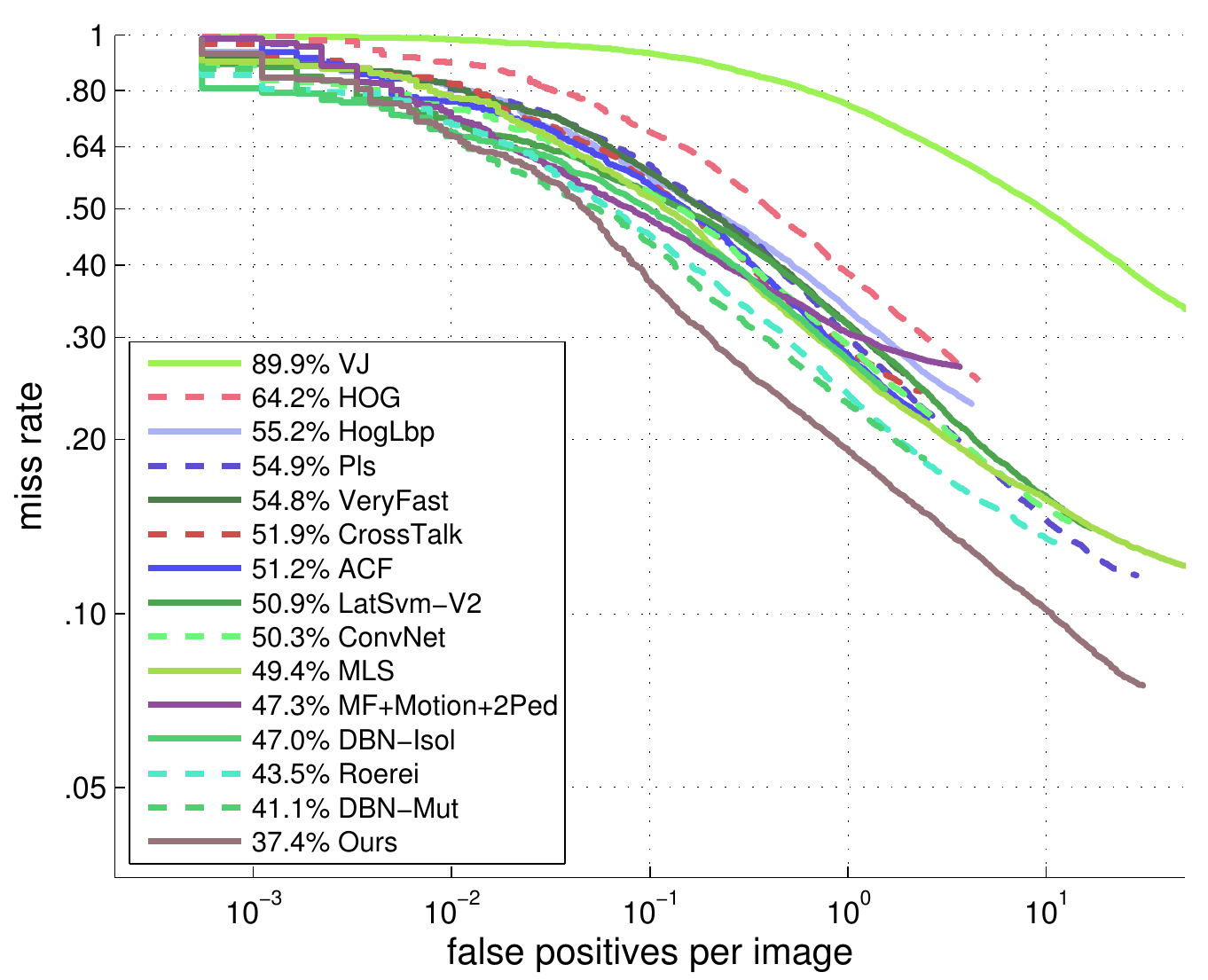}
        \includegraphics[width=0.46\textwidth,clip]{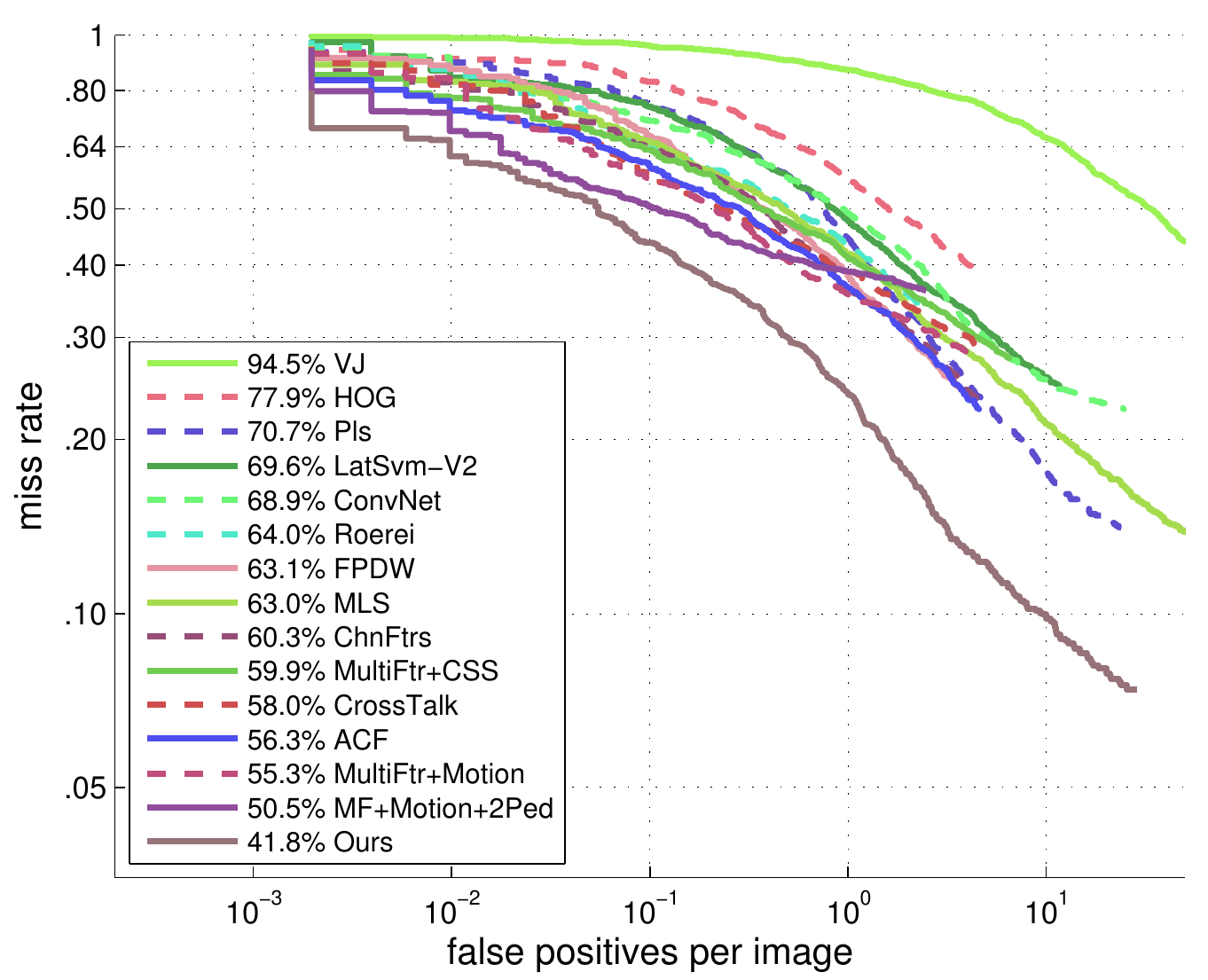}
        \includegraphics[width=0.46\textwidth,clip]{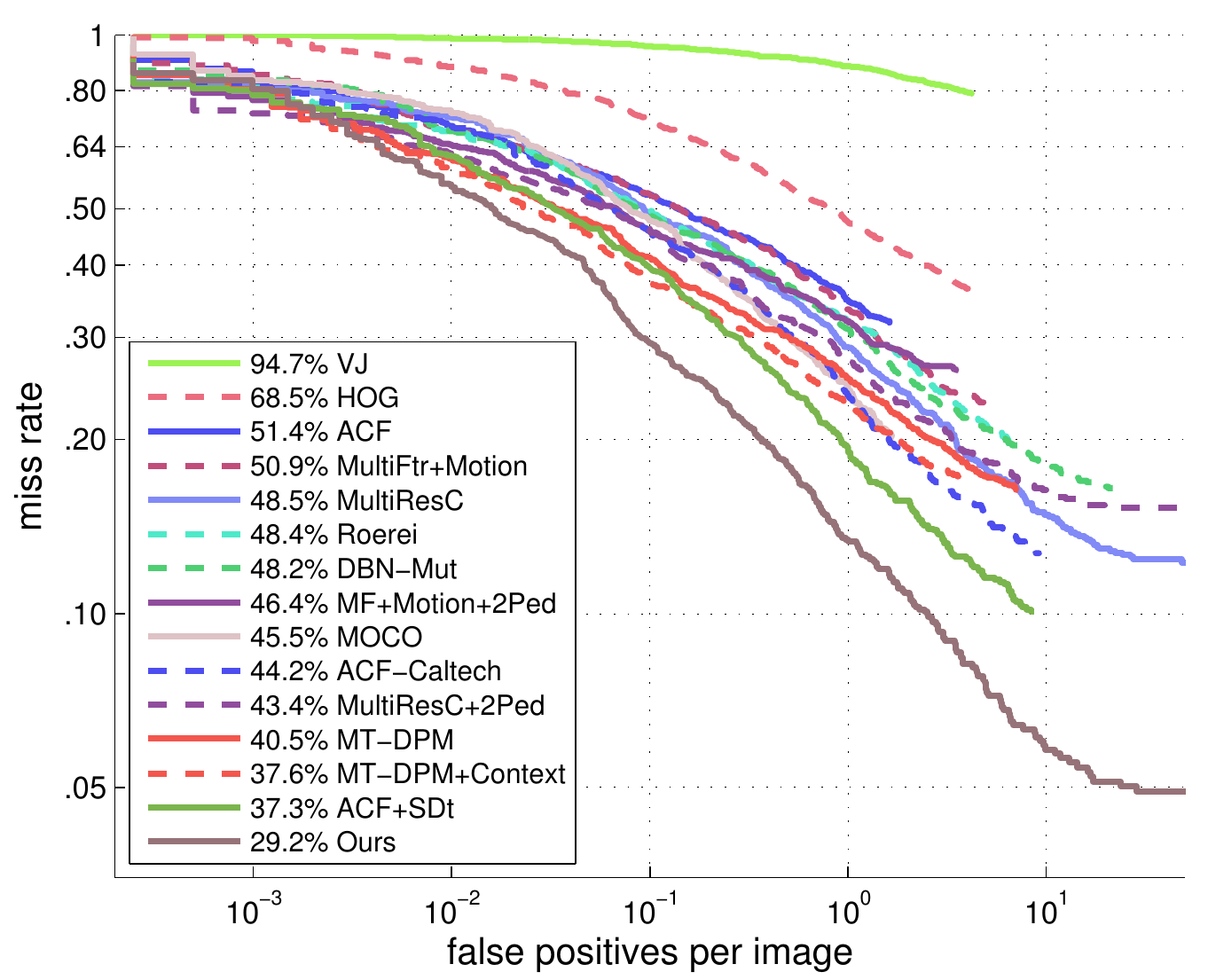}
    \caption{
    ROC curves of our proposed approach on INRIA, ETH, TUD-Brussels and Caltech-USA pedestrian detection benchmarks.
    }
    \label{fig:cov_fppi}
\end{figure*}

\begin{figure*}[t]
    \captionsetup[subfigure]{labelformat=empty}
    \centering
    \subfloat[Colour]{\includegraphics[width=0.14\textwidth,clip]{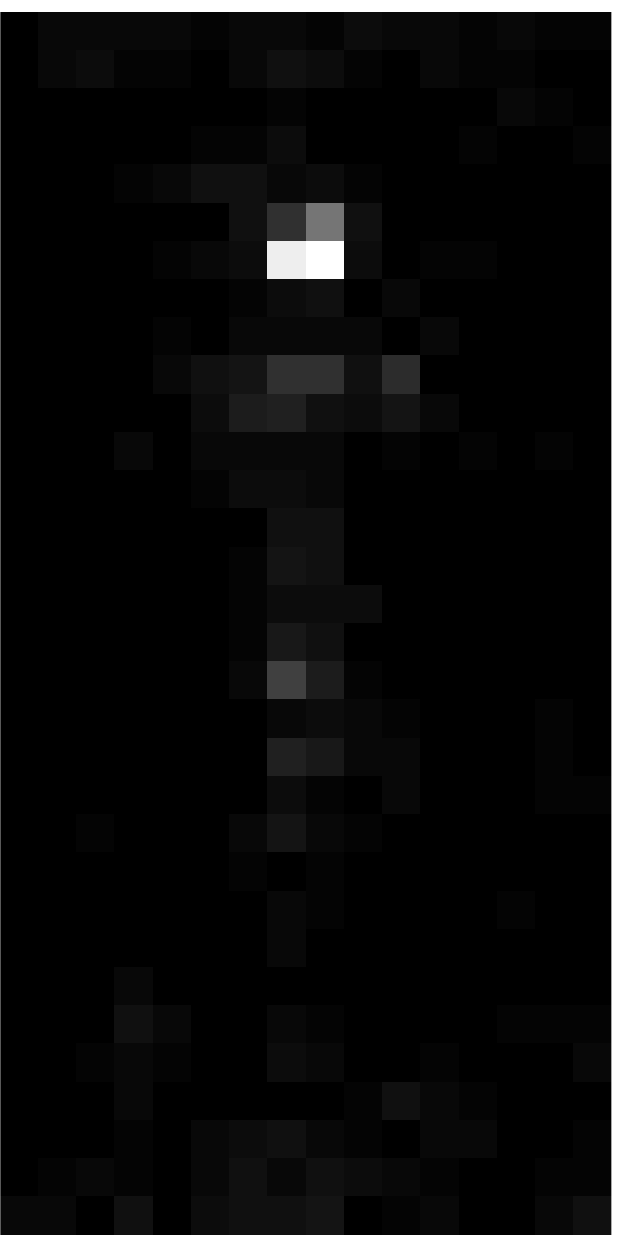}}
    ~
    \subfloat[Magnitude]{\includegraphics[width=0.14\textwidth,clip]{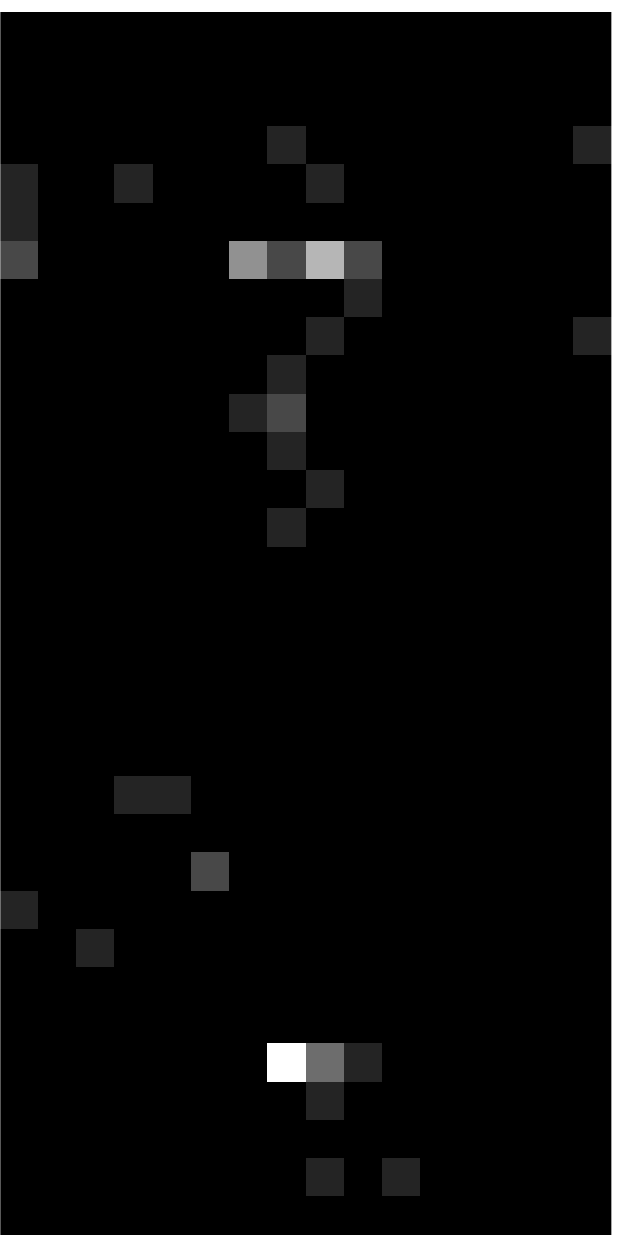}}
    ~
    \subfloat[Edges]{\includegraphics[width=0.14\textwidth,clip]{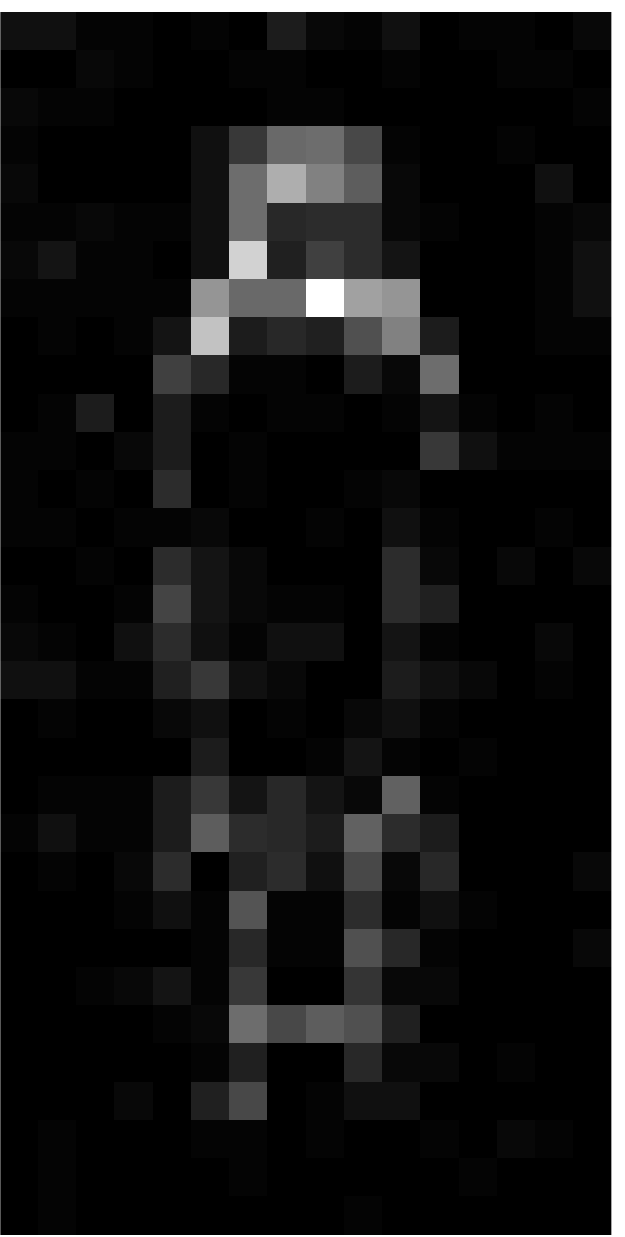}}
    ~
    \subfloat[\SLBP]{\includegraphics[width=0.14\textwidth,clip]{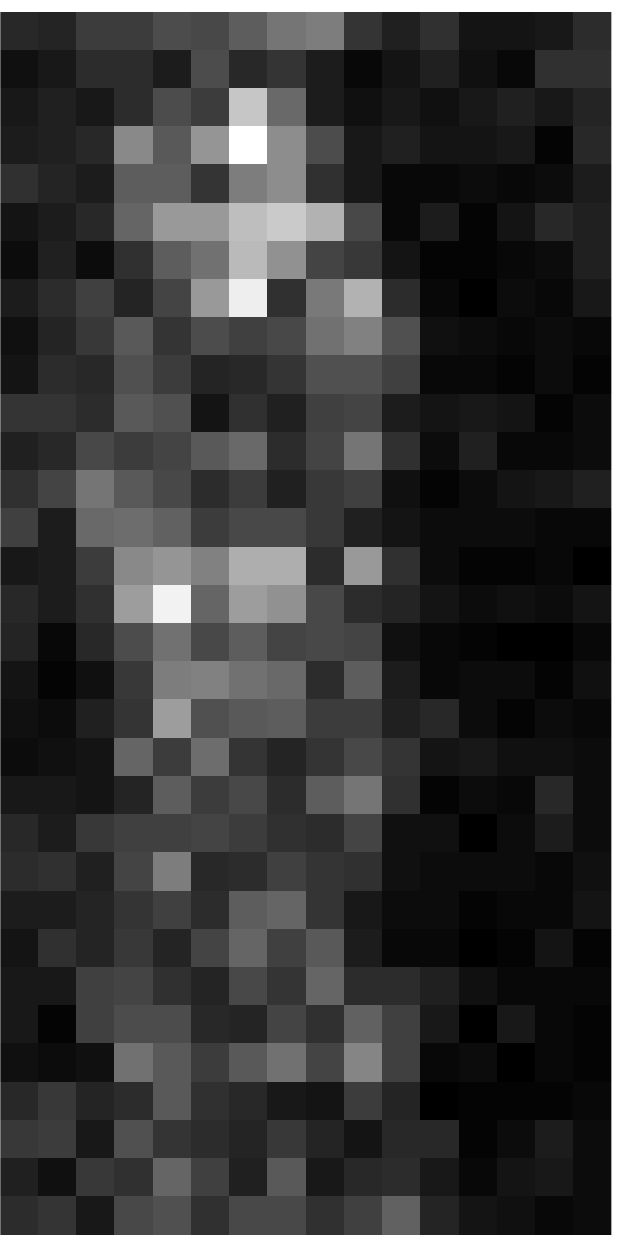}}
    ~
    \subfloat[\SCOV]{\includegraphics[width=0.14\textwidth,clip]{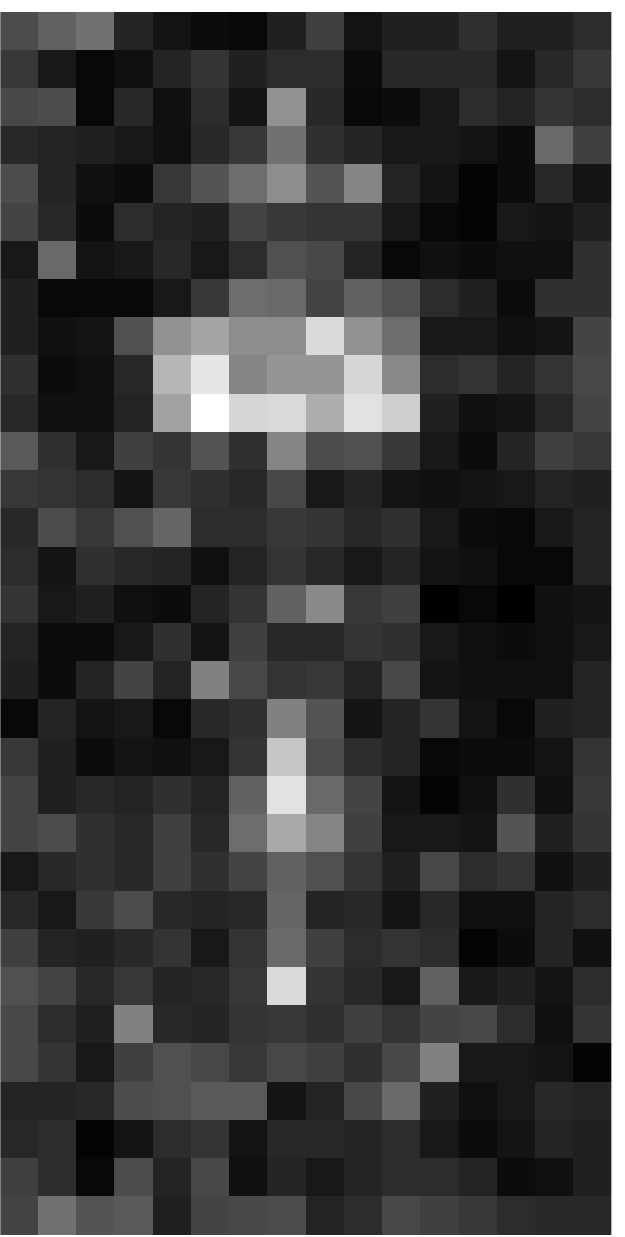}}
    ~
    \subfloat[ALL]{\includegraphics[width=0.14\textwidth,clip]{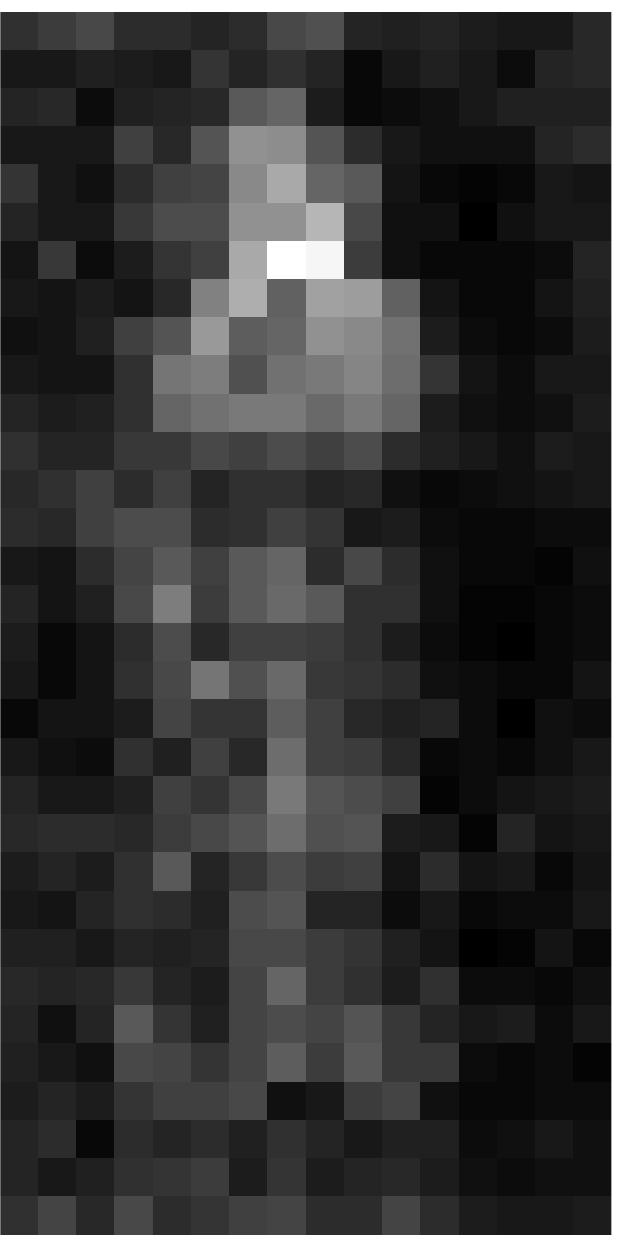}}
    \caption{
    Spatial distribution of selected regions based on their feature types.
    White pixels indicate that a large number of features are selected in that area.
    Often selected regions correspond to human contour and human body.
    }
    \label{fig:spatial_dist}
\end{figure*}

\subsection{Comparison with state-of-the-art results}
In the next experiment, we compare our combined features with state-of-the-art detectors.
Recently \cite{Lim2013Sketch} propose sketch tokens (ST) feature which achieves
the state-of-the-art result on the INRIA test set (a miss rate of $13.3\%$).
Our new detector outperforms ST by achieving a miss rate of $11.2\%$.
Our best performance is achieved when we apply \paucstruct to the combined features
(a miss rate of $11.1\%$).
As shown in Table~\ref{tab:tab1}, the combined features + \paucstruct outperform
all previous best results on four
major pedestrian detection benchmarks.
Fig.~\ref{fig:cov_fppi} compares our best results (the last row in Table~\ref{tab:tab1})
with other state-of-the-art methods.
Fig.~\ref{fig:spatial_dist} shows the spatial distribution of regions selected
by different feature types.
White pixels indicate that a large number of features are selected
in that region.
From the figure,
most selected regions typically contain
human contours (especially the head and shoulders).
Colour features are selected around the human face (skin colour)
while edge features are mainly selected around human contours
(head, shoulders and feet).
\SLBP features are selected near human head and human hips
while \SCOV features are selected around human chest and regions
between two human legs.

It is important to point out that our
significant improvement comes at the cost of increased computational complexity.
We briefly list these additional computational costs compared to \cite{Dollar2014Fast} here.
(i) Additional CPU time to extract two additional features:
\SCOV and \SLBP.
(ii) The time taken to re-compute the confidence score of positive patches.
To be more specific, we additionally calculate the dot product of the weak learners' output
and \paucstruct variables (new coefficients for weak learners), \ie,
$\vw^\T \vh$ where
$\vw$ is \paucstruct variables,
$\vh = [h_1(\cdot), \cdots, h_t(\cdot)]$
and $h_k(\cdot)$ is the $k$-th weak learner.
(iii) Additional CPU time to perform the global normalization (ACE).
In our experiment, applying the colour normalization on a $640 \times 480$
pixels image takes approximately $0.3$ seconds.
This fast result is already based on an approximation of ACE \cite{Getreuer2012Automatic}, which
estimates a slope function with an odd polynomial approximation
and uses the DCT transform to speed up the convolutions.
Using a single core Intel Xeon CPU $2.70$GHz processor,
our detector currently operates at approximately $0.126$ frames per second
(without global normalization) and $0.119$ frames per second (with global normalization)
on the Caltech data sets (detecting pedestrians larger than $50$ pixels).

\section{Conclusion}
In this paper we propose a simple yet effective feature extraction method
based on spatially pooled low-level visual features.
To achieve  optimal log-average miss rate performance measure,
we learn another set of weak learners' coefficients
whose aim is to improve the detection rate at the range of most practical importance.
The combination of our approaches contributes to a pedestrian detector which
outperforms all competitors on all of the standard benchmark datasets.
Based on our experiments, we observe that the choice of discriminative features and
implementation details are crucial to achieve the best detection performance.
Future work includes incorporating motion information through the use of
spatial and temporal pooling to further improve the detection performance.

%\paragraph
\textbf{Acknowledgements}
%
% The authors thank Dr. Rodrigo Benenson for his helpful suggestions and comments.
This work was in part supported by Australian Research Council grant FT120100969.

{%
\bibliographystyle{splncs}
\bibliography{draft}
}

\end{document}